\documentclass[runningheads,a4paper]{llncs}
\usepackage{llncsdoc}
\usepackage[pdftex]{graphicx}
\usepackage[caption=false,font=footnotesize]{subfig}
\usepackage{authblk}
\usepackage{url}
\usepackage[misc]{ifsym}
\usepackage{comment}
\usepackage{url}
\urldef{\mailsa}\path|{khpaeng, shwang, sgpark, mskim}@lunit.io|
\newcommand{\keywords}[1]{\par\addvspace\baselineskip
\noindent\keywordname\enspace\ignorespaces#1}
\begin{document}
\mainmatter  
\title{A Unified Framework for Tumor Proliferation Score Prediction in Breast Histopathology}

 \titlerunning{A Unified Framework for Tumor Proliferation Score Prediction}
\author{Kyunghyun Paeng\and Sangheum Hwang\and Sunggyun Park\and Minsoo Kim}
\authorrunning{K. Paeng et al.}
\institute{Lunit Inc., Seoul, Korea\\
\mailsa}

\toctitle{Lecture Notes in Computer Science}
\tocauthor{Authors' Instructions}
\maketitle

\begin{abstract}
We present a unified framework to predict tumor proliferation scores from breast histopathology whole slide images. 
Our system offers a fully automated solution to predicting both a molecular data-based, and a mitosis counting-based tumor proliferation score. 
The framework integrates three modules, each fine-tuned to maximize the overall performance: An image processing component for handling whole slide images, a deep learning based mitosis detection network, and a proliferation scores prediction module. 
We have achieved 0.567 quadratic weighted Cohen's kappa in mitosis counting-based score prediction and 0.652 F1-score in mitosis detection. 
On Spearman's correlation coefficient, which evaluates predictive accuracy on the molecular data based score, the system obtained 0.6171. 
Our approach won first place in all of the three tasks in Tumor Proliferation Assessment Challenge 2016 which is MICCAI grand challenge.
\end{abstract}

\section{Introduction}
Tumor proliferation speed is an important biomarker for estimating the prognosis of breast cancer patients~\cite{van2004prognostic}.
The mitotic count is part of the Bloom \& Richardson grading system~\cite{elston1991pathological}, and a well-recognized prognostic factor~\cite{baak2005prospective}.
However, the procedure currently used by pathologists to count the number of mitoses is tedious and subjective, and can suffer from reproducibility problems~\cite{veta2016mitosis}.
Automatic methods have rapidly advanced the state-of-the-art in mitosis detection ~\cite{albarqouni2016aggnet,cirecsan2013mitosis,veta2015assessment}.
In particular, systems based on deep convolutional neural networks(DCNN) have been successfully adopted to mitosis detection~\cite{cirecsan2013mitosis}. 
The latter family of algorithms have advanced the performance of automatic mitosis detection to near-human levels, offering promise in addressing the problem of subjectivity and reproducibility~\cite{veta2016mitosis}.

In a practical scenario, it is desirable to provide tumor proliferation assessment at the whole slide image (WSI) level. 
The size of the WSI is exceedingly large, posing additional challenges for applying automatic techniques. 
Previous studies carried out mitosis detection within ROI patches from the WSI which were pre-selected by a pathologist. 
On the contrary, automatic analysis of the WSI must resolve the issue of ROI selection without human guidance. 
An additional module is required to perform the final mapping between a mitosis counting result and the tumor proliferation score.

In this paper, we present our work in addressing each of the issues outlined above, in the form of a unified framework for predicting a tumor proliferation score directly from breast histopathology WSIs. 
Our pipeline system performs fully automatic prediction of tumor proliferation scores from WSIs. 
Our system participated in a well-recognized challenge and was able to outperform other systems in each subtask, as well as in the final proliferation score prediction task, validating the design choices of each module, as well as the capability of the integrated system.

\section{Methodology}

\begin{figure}[t]
\begin{center}
   \includegraphics[width=0.8\linewidth]{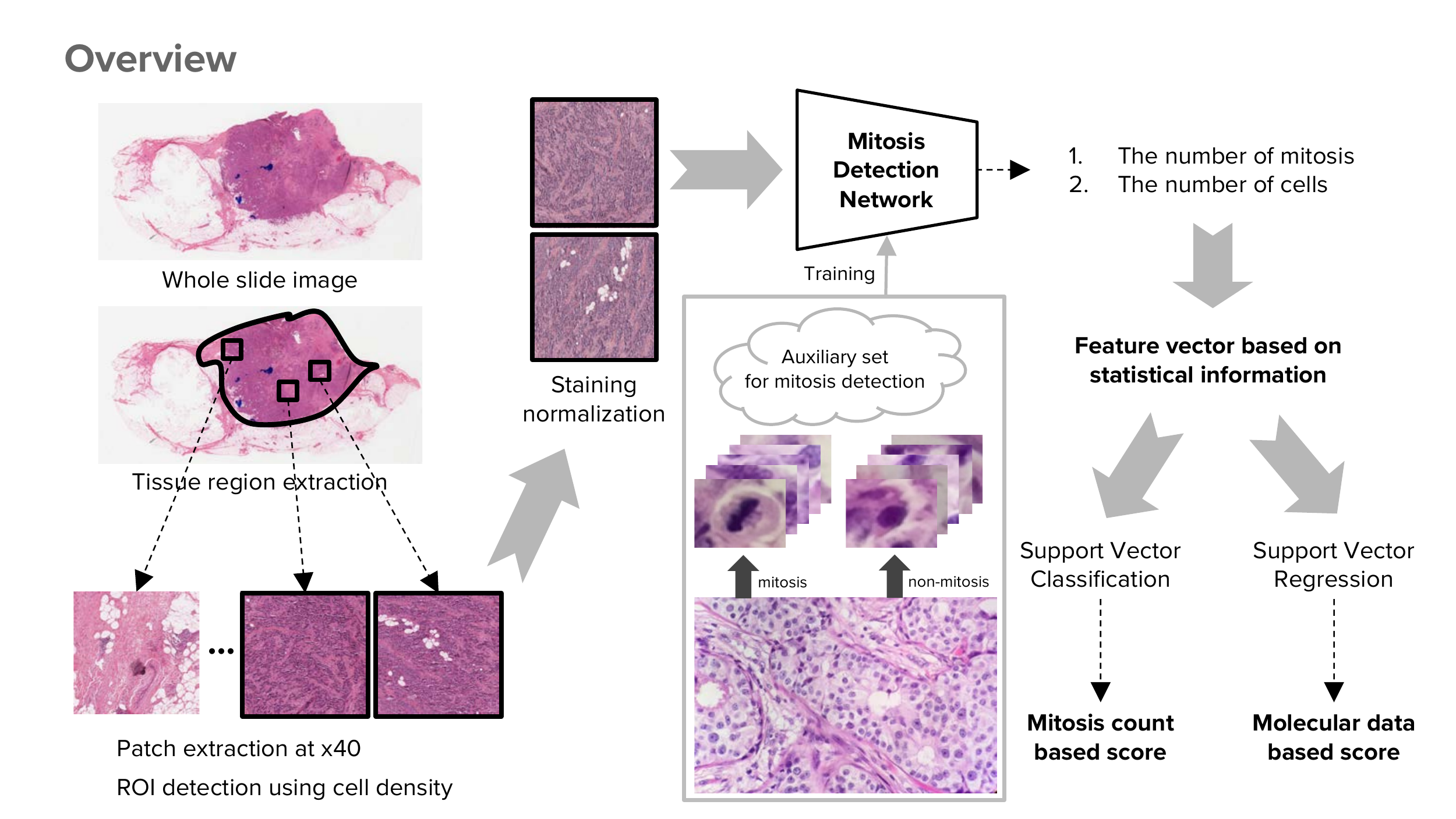}
\end{center}
   \caption{Tumor proliferation score prediction framework. First, tissue region and patch extraction is performed and ROI patches are selected. After staining normalization, we count the number of mitoses using a DCNN based detection network. The mitosis detection network is trained by using a dataset containing annotated mitosis regions. The results of mitosis detection are converted to a feature vector in each WSI. Finally, the tumor proliferation scores are predicted by SVMs. }
\label{fig:overview}
\end{figure}

\subsection{Whole Slide Image Handling}
\begin{figure}[t]
\begin{center}
   \includegraphics[width=\linewidth]{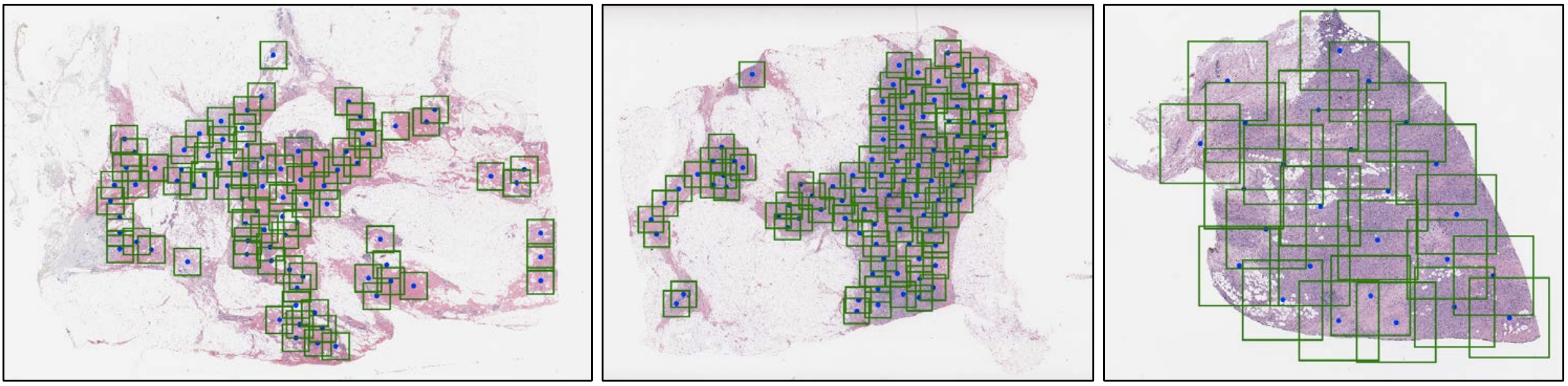}
\end{center}
   \caption{The results of tissue region and patch extraction. The blue dots indicate the position of the sampling point (i.e., the centers of sampled patches). Each green square shows a 10 consecutive HPFs region.}
\label{fig:tissue}
\end{figure}

{\bf Tissue region and patch extraction:}
We first identify tissue blobs from WSI, through a combination of WSI resizing, tissue region extraction by Otsu's method, and binary dilation.
Once the tissue blobs have been determined, we sample patches from them to be used in mitosis counting.
There are many ways to perform the patch sampling, in terms of the choice of patch size and sampling step.
Recalling the fact that Bloom \& Richardson (B\&R) grading depends on the number of mitoses within a 10 consecutive high power fields (HPFs) region, an area of approximately 2 mm$^2$, we extract patches corresponding to 10 HPFs.
However, the definition of 10 consecutive HPFs region is highly variant, for instance, it may have only horizontal or vertical directions. 
For generality, we define 10 consecutive HPFs to be a square.
The average number of patches extracted in each slide is roughly 300.
The result of tissue region and patch extraction is shown in Fig.~\ref{fig:tissue}.

\noindent
{\bf Region of interests detection:}
To select region of interests (ROIs) from the sampled patches, we utilize the fact that there are generally many mitoses in regions with high cell density.
We use CellProfiler~\cite{carpenter2006cellprofiler,kamentsky2011improved}, to estimate the cell density of the patches.
Finally, we obtain the number of cells in each patch, and after sorting patches from one WSI according to the number of cells, the top $K$ patches are chosen as ROIs.
We set $K$ to 30 with consideration for computation cost.

\noindent
{\bf Staining normalization:}
Because WSIs vary highly in appearance, it is beneficial to normalize the staining quality of ROI patches.
In order to normalize the staining of patches, we apply the method described in~\cite{macenko2009method}.
We set $\alpha$ and $\beta$ to 1.0 and 0.15 respectively.

\subsection{Deep Convolutional Neural Networks based Mitosis Detection}
{\bf Training procedure:}
After selecting ROI patches, we detect mitoses using the trained model.
The mitosis detector is trained on 128x128 patches extracted from the auxiliary dataset.
We use a two step training procedure in order to reduce the false positive rate~\cite{cirecsan2013mitosis}.
First, a first-pass training of the network is performed on an initial dataset.
Then, a list of image regions that the network has identified as false positive mitoses cases is extracted by the trained network.
After that, we build a new, second training dataset which consists of the same ground truth mitosis samples and normal samples from before, but with an additional 100,000 normal samples (i.e., false positives) generated with random translation augmentation from the initially trained network.
In total, the new dataset consists of 70,000 mitosis patches and 280,000 normal patches.
We retrain the network from scratch on this new dataset to obtain the final mitosis detection model.

\begin{figure}[t]
\begin{center}
   \includegraphics[width=0.8\linewidth]{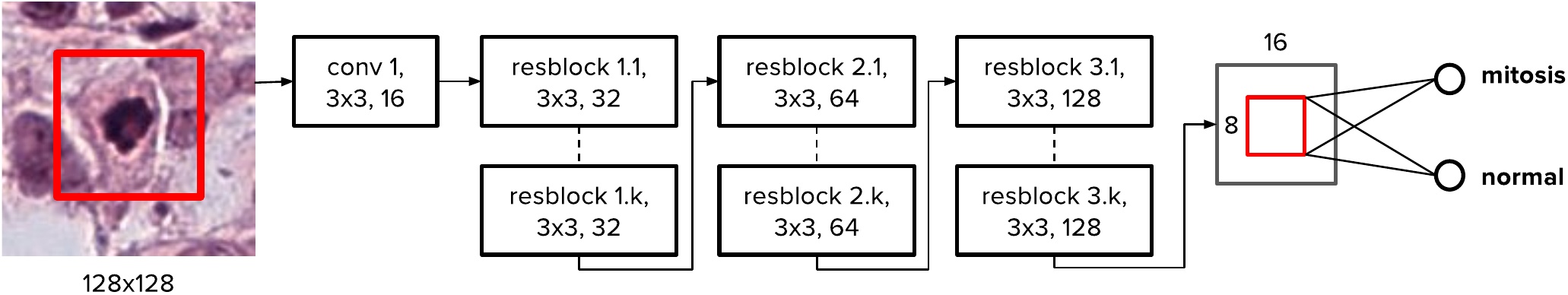}
\end{center}
   \caption{The {\it L-view} network architecture for mitosis detection. $k$ is a depth factor (2 or 3). The dotted line indicates that it is repeated $k$ times. Two red rectangles indicate corresponding regions in the architecture.}
\label{fig:mts_arch}
\end{figure}

\begin{figure}[b]
\begin{center}
   \includegraphics[width=0.7\linewidth]{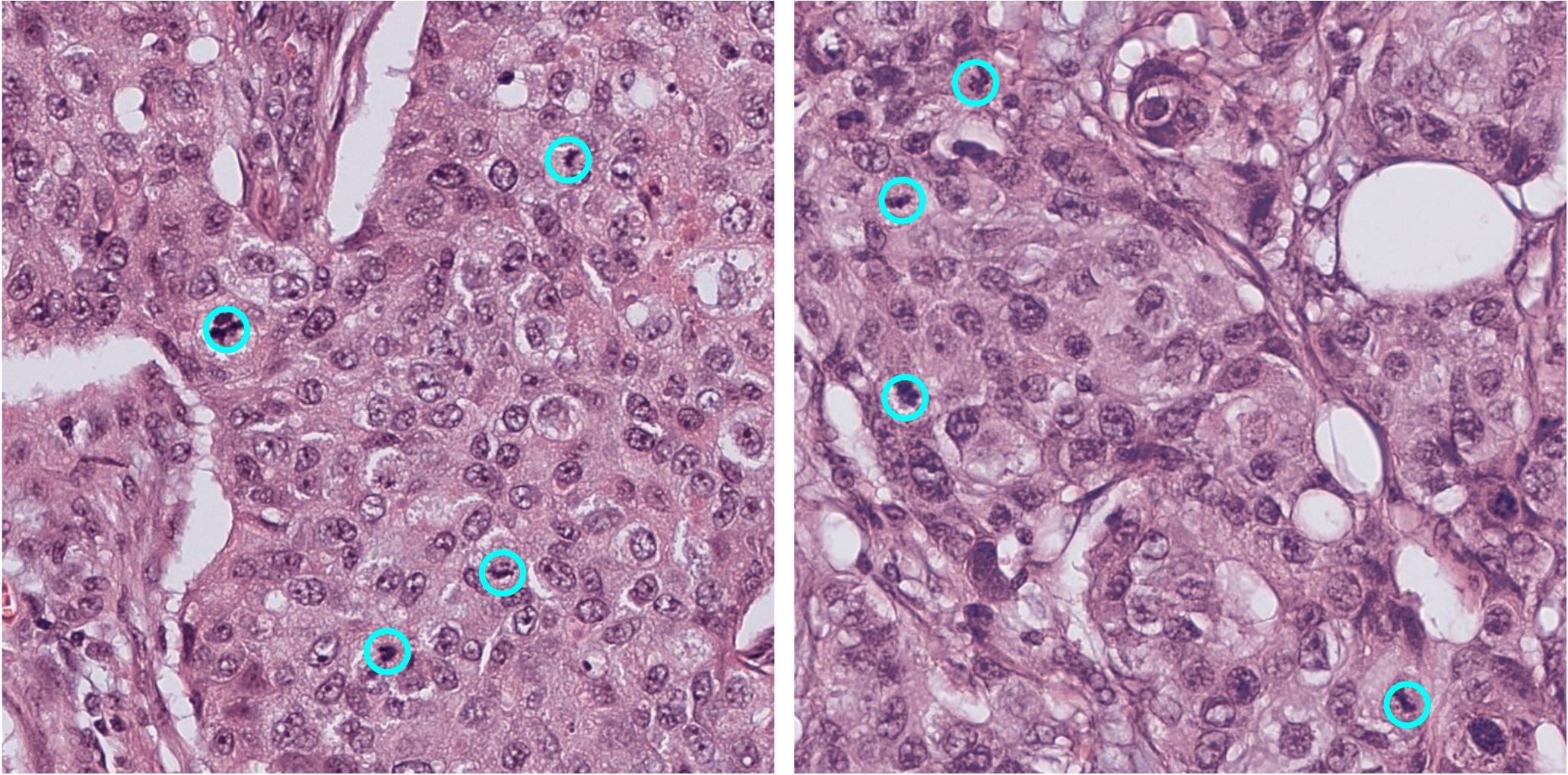}
\end{center}
   \caption{Mitosis detection examples. The detected mitoses are highlighted with cyan circles.}
\label{fig:mts}
\end{figure}

\noindent
{\bf Architecture:}
The architecture of the mitosis detection network is based on the Residual Network (ResNet)~\cite{he2016identity}, and it consists of 6 or 9 residual blocks.
We train a network on 128x128 input patches, then convert the trained network to a fully convolutional network~\cite{long2015fully}, which can then be applied directly to an entire ROI patch for inference.
This method offers much greater computational efficiency over the alternative of subdividing the ROI patch and performing piecewise inference.
However, although the fully convolutional approach is much faster, it fails to exactly match the performance of the standard subdivision approach.
This problem arises from the large discrepancy between the input size of the network during training and inference when using the fully convolutional technique, exacerbated by the ResNet's large depth leading to a larger receptive field and corresponding larger zero-padding exposure. 
In order to alleviate this problem, we introduce a novel architecture named large-view ({\it L-view}) model.
Fig.~\ref{fig:mts_arch} shows the {\it L-view} architecture.
Although the {\it L-view} architecture has 128x128 input size, in the final global pooling layer, we only activate a smaller region corresponding to the central 64 x 64 region in the input patch.
This allows the zero padding region to be ignored in the training phase and consequently solves the problem of the zero padding effect, leading to a sizable improvement in mitosis detection score.
Fig.~\ref{fig:mts} represents the examples of mitosis detection results.

\subsection{Tumor Proliferation Score Prediction}
In the final module, we encode each slide into a feature vector and train an SVM to predict the tumor proliferation scores.
The feature vector consists of the number of mitoses and cells in each patch through the steps previously described, in addition to 21 features based on statistical information, as shown in TABLE~\ref{table:feat}.
Since we have no prior knowledge of which features are relevant for tumor proliferation score prediction, we find the best feature vector through cross validation from various feature combinations. We performed 10-fold cross validation using 500 training pairs with various $C$ values of support vector machines (SVMs) with radial basis function (RBF) kernel in which the gamma of RBF kernel is fixed to 1/the dimension of feature. Finally, we selected combinations of features with the best performance. To reduce the search permutations, we reduced the dimension of the feature vector to 14 features.

\begin{table}[t]
\caption{Feature components for predicting tumor proliferation scores. \# and mts represents the number and mitosis, respectively.
B\&R is Bloom\& Richardson grading. Top 10\% and 30-70\% means rank 1-3 and rank 9-21 from sorted ROIs.
min, avg, and max indicate the minimum, average, and maximum value in selected part from ROIs.}
\label{table:feat}
\centering
\begin{tabular}{|c|c|c|c|c|c|}
\hline
\bf{No.} & \bf{feat. description} & \bf{No.} & \bf{feat. description} & \bf{No.} & \bf{feat. description} \\
\hline
\hline
0 & avg. \# mts & 7 & std. \# cells & 14 & std. \# mts in top 10\% \\
1 & max. \# mts & 8 & ratio of avg. mts/cells & 15 & std. \# cells in top 10\% \\
2 & std. \# mts & 9 & ratio of max. mts/cells & 16 & min. \# mts in top 10\% \\
3 & B\&R of avg. \# mts & 10 & avg. \# mts in top 10\% & 17 & min. \# cells in top 10\% \\
4 & B\&R of max. \# mts & 11 & min. \# mts & 18 & avg. \# mts in 30-70\% \\
5 & avg. \# cells & 12 & min. \# cells & 19 & max. \# mts in 30-70\% \\
6 & max. \# cells & 13 & ratio of min. mts/cells & 20 & std. \# mts in 30-70\% \\
\hline
\end{tabular}
\end{table}

\section{Results}
In this section, we introduce the dataset used for training and validation and show the performance of each individual component in the proposed system. 
In addition, all experimental results except for validation were evaluated in a public challenge.

\subsection{Datasets}
We used two datasets\footnote{http://tupac.tue-image.nl/node/3} for developing the fully automated tumor proliferation score prediction system.
First, the auxiliary dataset containing annotated mitosis regions from two pathologists are used to build a mitosis detector.
The training dataset includes 656 patches of 1 HPF or 10 HPFs from 73 patients from three pathology centers, and 34 patches of 10 HPFs from 34 patients are used for testing.
The training dataset is randomly split to validate the mitosis detector, and the validation dataset includes 142 mitoses from 6 patients.
Secondly, to train the entire system, we used the main dataset which consists of 500 training WSIs and 321 testing WSIs.
Each WSI has two corresponding scores, a mitosis counting based score indicating one of three classes and a molecular data based score~\cite{nielsen2010comparison} which is a continuous value. To validate the entire system, we used 10-fold cross validation.

\subsection{Experiments}
We evaluated on three tasks to validate the performance of the proposed system.
Our results were compared with the other approaches in the challenge.~\footnote{Compared methods are denoted by alphabet to anonymize the names of the participating teams.}
First, the mitosis detection performance is measured via F1-score.
The performance of the entire system is based on two evaluation metrics: quadratic weighted Cohen's kappa for the mitosis counting based score and Spearman's correlation coefficient for the molecular based score.

\noindent
{\bf Mitosis detection:}
We fixed the batch size to 128 and initial learning rate to 0.1, and applied the same two-step training procedure for all model configurations during training.
The $k$ of our {\it L-view} architecture is 3 in the first training phase, and color, brightness and contrast augmentation is used in various combinations.
The learning rate is 0.1 for the first 8 epochs of the first training phase, and 0.1, 0.01 and 0.001 for 8, 12 and 14 epochs of the second training phase.
The evaluation results are shown in TABLE~\ref{table:mts_res}.
With the {\it L-view} architecture, we obtain 0.731 and 0.652 F1-score on the validation and test, respectively.
In case of training without false positives, the performance degradation of 0.1 F1-score was observed in our validation set.

\begin{table}[t]
\caption{F1 scores in mitosis detection. * indicates that additional data was used for training. Proposed method is based on L-view architecture with 128x128 input size.}
\label{table:mts_res}
\centering
\begin{tabular}{|c|c|c|c|}
\hline
\bf{Method} & \bf{Validation} & \bf{Test} \\
\hline
\hline
Proposed & 0.731 & 0.652 \\
Team A$^*$ & - & 0.648 \\
Team B & - & 0.616 \\
Team C$^*$ & - & 0.601 \\
\hline
\end{tabular}
\end{table}

\noindent
{\bf Mitosis counting based proliferation score prediction:}
After selecting the mitosis detector with the best performance, we trained the tumor proliferation score prediction module.
We evaluated quadratic weighted Cohen's kappa score using 10-fold cross validation and found the best feature combination.
The min related feature values were found to be unimportant, so features 11, 12, 13, 16, 17 were removed from the combination list.
Finally, we found the best performance to be shown by a 12 dimensional feature vector of the features 0, 1, 2, 3, 4, 5, 6, 7, 10, 15, 18, 20 from TABLE~\ref{table:feat}. The $C$ value of SVMs was 0.03125.
The evaluation results are shown in TABLE~\ref{table:kappa}.
The kappa score was 0.504 on the validation dataset and 0.567 on the test dataset, and the proposed system outperformed all other approaches. In addition, our system is even better than both the semi-automatic model and the model trained by using additional data.

\begin{table}[t]
\caption{Mitosis counting based proliferation score prediction results in validation and test. The evaluation metric is Quadratic weighted Cohen's kappa score. * indicates that additional data was used for training. ** is the semi-automatic method where ROIs are selected by a pathologist.}
\label{table:kappa}
\centering
\begin{tabular}{|c|c|c|}
\hline
\bf{Method} & \bf{Validation} & \bf{Test} \\
\hline
\hline
Proposed & 0.504 & 0.567 \\
Team D$^{**}$ & - & 0.543 \\
Team B & - & 0.534 \\
Team E$^*$ & - & 0.462 \\
\hline
\end{tabular}
\end{table}

\begin{table}[b]
\caption{Molecular data based proliferation score prediction results in validation and test. Evaluation metric is Spearman’s correlation coefficient.}
\label{table:corr}
\centering
\begin{tabular}{|c|c|c|}
\hline
\bf{Method} & \bf{Validation} & \bf{Test} \\
\hline
\hline
Proposed & 0.642 & 0.617 \\
Team F & - & 0.516 \\
Team B & - & 0.503 \\
\hline
\end{tabular}
\end{table}

\noindent
{\bf Molecular data based proliferation score prediction:}
The same method is used to find the best feature combination for predicting the molecular data based score.
The min related features were once again found to be unimportant and removed.
The evaluation results are shown in TABLE~\ref{table:corr}.
The 13 dimensional best feature vector includes features 0, 1, 2, 3, 4, 5, 6, 7, 8, 10, 14, 18, 20 from TABLE~\ref{table:feat}, and the $C$ value of SVMs was 0.25.
We obtained 0.642 Spearman score on the validation dataset and 0.617 on the test dataset.
Our system significantly outperformed all other approaches.
Through the best feature combination, we discovered that the values of features 8 and 14, rather than 15, are important in predicting the molecular data based score.
In other words, the key factor for estimating this score is not cell information but the number of mitoses. 
This affirms the high correlation of mitotic count and tumor proliferation speed. 

\section{Conclusion}
We presented a fully automated unified system for predicting tumor proliferation scores directly from breast histopathology WSIs.
The proposed system enables fully automated modular prediction of two tumor proliferation scores based on mitosis counting and molecular data.
Our work confirms that a mitosis detection module could be integrated in prognostic grading system that is more practical in a clinical scenario.
In addition, we demonstrated that our system achieved state-of-the-art performance in proliferation scores prediction.

\bibliographystyle{splncs03}
\bibliography{./refer}

\end{document}